\documentclass[10pt,twocolumn,letterpaper]{article}

\usepackage[pagenumbers]{cvpr} 

\definecolor{cvprblue}{rgb}{0.21,0.49,0.74}
\usepackage[pagebackref,breaklinks,colorlinks,allcolors=cvprblue]{hyperref}

\usepackage{graphicx}
\usepackage{tabularx,booktabs}
\usepackage{svg}
\newcolumntype{C}{>{\centering\arraybackslash}X} 
\usepackage{textcomp}
\usepackage{xcolor}
\usepackage{siunitx}
\usepackage[export]{adjustbox}
\usepackage{makecell}

\def\paperID{4} 
\def\confName{CVPR WAD}
\def\confYear{2025}

\title{Inferring Driving Maps by Deep Learning-based Trail Map Extraction}

\author{Michael Hubbertz\textsuperscript{1,2}\\
\and
Pascal Colling\textsuperscript{2}\\
\and
Qi Han\textsuperscript{2}\\
\and
Tobias Meisen\textsuperscript{1}\\
\and
    \centerline{
        \textsuperscript{1}\textit{University of Wuppertal}, {\tt <lastname>@uni-wuppertal.de}
    } \\
    \centerline{
        \textsuperscript{2}\textit{Aptiv Services Deutschland GmbH}, {\tt <firstname>.<lastname>@aptiv.de}
    }
}

\begin{document}
\maketitle
\begin{abstract}
High-definition (HD) maps offer extensive and accurate environmental information about the driving scene, making them a crucial and essential element for planning within autonomous driving systems. To avoid extensive efforts from manual labeling, methods for automating the map creation have emerged. Recent trends have moved from offline mapping to online mapping, ensuring availability and actuality of the utilized maps. While the performance has increased in recent years, online mapping still faces challenges regarding temporal consistency, sensor occlusion, runtime, and generalization. We propose a novel offline mapping approach that integrates trails — informal routes used by drivers — into the map creation process. Our method aggregates trail data from the ego vehicle and other traffic participants to construct a comprehensive global map using transformer-based deep learning models. Unlike traditional offline mapping, our approach enables continuous updates while remaining sensor-agnostic, facilitating efficient data transfer. Our method demonstrates superior performance compared to state-of-the-art online mapping approaches, achieving improved generalization to previously unseen environments and sensor configurations. We validate our approach on two benchmark datasets, highlighting its robustness and applicability in autonomous driving systems.
\end{abstract}    
\section{Introduction}

Maps are fundamental tools for navigation, providing essential spatial information that enables drivers to traverse complex road networks safely and efficiently. While standard definition (SD) maps are sufficient for human drivers, offering a general overview of road layouts, high-definition (HD) maps are becoming indispensable for autonomous vehicles. These maps provide fine-grained details, such as lane geometries and road surface attributes, which complement onboard sensor perception of the dynamic environment and ensure precise localization and safe navigation \cite{zhibin_bao_high-definition_2022}.

\begin{figure}[t] 
    \centering
    \includegraphics[width=0.40\textwidth]{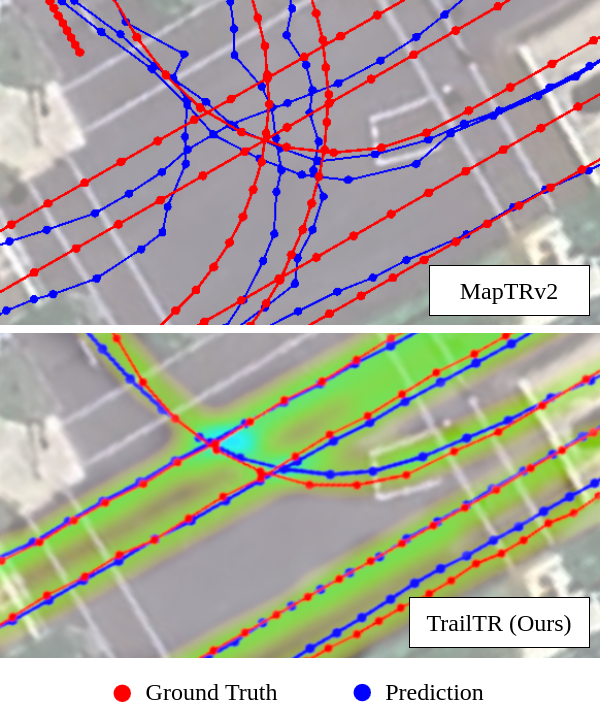}\label{fig:eye_catcher}
    \caption{Visualization of the ground truth lane centerlines and predictions of a popular online mapping model (MapTRv2) and our approach (TrailTR) for a nuScenes validation sample. Both models were trained on the geographically disjoint near extrapolation split \cite{lilja_localization_2024}. Centerlines with no corresponding trail data were removed from the ground truth in our approach.}
\end{figure}

Traditionally, HD maps have been created through labor-intensive processes, including manual labeling and correction. However, these methods lack scalability, making automated HD map generation a critical research focus \cite{jiao_machine_2018}. Thereby, two primary approaches have emerged: offline mapping and online mapping \cite{zhibin_bao_high-definition_2022}. 

Offline mapping relies on pre-recorded data, such as aerial imagery or prior sensor observations, to construct a globally referenced, highly detailed map. While these maps offer high accuracy, they require continuous updates to account for road modifications, such as construction zones \cite{pannen_how_2020}. Furthermore, offline HD maps are limited to predefined regions, preventing autonomous vehicles that rely on map data from operating in unmapped areas.


To address these limitations, online mapping has gained increasing attention. Online approaches generate local, real-time maps by leveraging onboard sensor data, including camera and Lidar inputs \cite{li_hdmapnet_2022, liao_maptr_2023, liu_vectormapnet_2023, liao_maptrv2_2023}. 


However, the created map based on current online sensor data can suffer from occlusion in certain scenarios, resulting in maps that are not completely covering the roads of the traffic situation. Recent research has also shown limitations regarding generalization to new areas of online mapping models \cite{lilja_localization_2024}. 
Furthermore, online mapping approaches, which are often transformer-based, require high computational power for real-time execution \cite{zhong_transformer-based_2023}, which is challenging due to the limited processing capacity of production vehicles. Additionally, high-resolution Lidar-based solutions, while providing detailed 3D representations, remain too costly for widespread adoption and are mainly used in research.


A promising approach to improving map creation is the integration of driving trails — frequently traveled routes derived from real-world traffic patterns. These trails capture informal driving behaviors that may not be easily inferred from static lane markings, providing valuable insights into human-like navigation strategies. By incorporating this information, autonomous vehicles can follow natural, behaviorally aligned paths, enhancing adaptability in diverse environments. This integration enables a more flexible and context-aware driving strategy, bridging the gap between conventional lane-based navigation and real-world driving behavior. \cite{ding_flowmap_2023}

In this paper, we propose an offline mapping approach that overcomes the classical limitations regarding coverage and required constant updates by incorporating trails paired with state-of-the-art transformer neural network architectures from the online mapping domain. Our method aggregates trajectory data from the ego vehicle and other traffic participants to construct a global HD map, leveraging deep learning to extract meaningful features from the collected trails. A key advantage of our approach is its sensor-agnostic nature — it does not rely on specific sensor configurations and can utilize data from any vehicle equipped with 3D object detection and global localization systems.  This enables scalability and facilitates deployment across large fleets of vehicles. Furthermore, downstream applications such as motion prediction and path planning benefit from the enriched lane structure information \cite{ulbrich_towards_2017, dauner_parting_2023}. Unlike conventional mapping techniques that focus solely on static lane features, our approach captures human-like driving patterns, allowing autonomous systems to better align their trajectories with real-world driving behaviors.


Our key contributions can be summarized as follows: 

\begin{itemize}
    \item We introduce a novel crowdsourced offline mapping framework that extracts high-definition directed driving maps from recorded trail data using a state-of-the-art transformer-based deep learning model.
    \item We convert discrete trail data into a neural network input representation by applying preprocessing and data augmentation strategies. Our method enhances feature extraction and improves the model's robustness and generalization across diverse environments.
    \item Our approach serves as an alternative to existing online mapping approaches by offering improved accuracy and generalization and allows tile-wise map updates.
\end{itemize}
\section{Related Work}



\noindent \textbf{Traditional offline map generation approaches}:{\enspace\;} In the early stages of offline mapping, approaches primarily relied on aggregating pre-recorded data such as point clouds, aerial imagery, and GPS trajectories  \cite{zhibin_bao_high-definition_2022}. The constructed maps depicted the high-level structure of the road network. In this context, centerline representations primarily focus on the general road topology, rather than lane-level structures. Consequently, these early maps lacked critical precision and contextual information required for high-accuracy autonomous navigation. 


\vspace{5mm} 

\noindent \textbf{Online HD map learning}:{\enspace\;}Online HD mapping approaches construct HD maps in real-time by processing onboard sensor data. Modern methods utilize a combination of sensor modalities, typically including single front-view camera images, camera surround views consisting of multiple images and a sensor point cloud from Radar or Lidar sensors. Due to the complexity of these multimodal inputs, recent research predominantly relies on machine learning-based techniques, as rule-based algorithms are insufficient for capturing the intricate spatial and semantic relationships required for HD map construction \cite{li_hdmapnet_2022, liao_maptr_2023, liu_vectormapnet_2023, liao_maptrv2_2023}. 


The first online mapping approach capable of generating vectorized map elements directly from sensor data is HDMapNet \cite{li_hdmapnet_2022}. Its architecture employs a combination of encoder-decoder networks to predict segmentation maps, instance embeddings, and directional cues, which are subsequently fused to construct a vectorized HD map.

More recent architectures in online mapping follow a similar structure, largely inspired by the Detection Transformer (DETR) framework \cite{carion_end--end_2020}. In these models, a feature extraction backbone processes the input and projects it into the bird’s-eye view (BEV) space. These features are then attended to by learnable object queries within a transformer decoder. The output of the decoder is then passed to a prediction head consisting of a classification and regression head. The former predicts the class of the map object, while the latter predicts its shape represented by discrete points on the map. Examples of this type of architecture can be seen in MapTR \cite{liao_maptr_2023}, VectorMapNet \cite{liu_vectormapnet_2023} and MapTRv2 \cite{liao_maptrv2_2023}, with the last two being able to also predict lane centerlines. Further approaches since then have optimized the architecture, e.g. through mask guiding the prediction of map objects \cite{liu_mgmap_2024}, improving the query design of the transformer decoder \cite{liu_leveraging_2024}, integrating additional information from global map priors \cite{xiong_neural_2023} or performing temporal fusion of the frames to overcome inconsistency limitations due to occlusion and complexity \cite{li_dtclmapper_2024, yuan_streammapnet_2024, chen_maptracker_2024, peng_prevpredmap_2024}.

In our approach, we leverage a similar DETR-based architecture, but we modify the input features to work with collected and processed detection data. This mitigates problems regarding occlusion and temporal inconsistency, while unifying the input representation and thereby enabling great generalization properties.

\begin{figure*}[t] 
\centering
\includegraphics[width=0.95\textwidth]{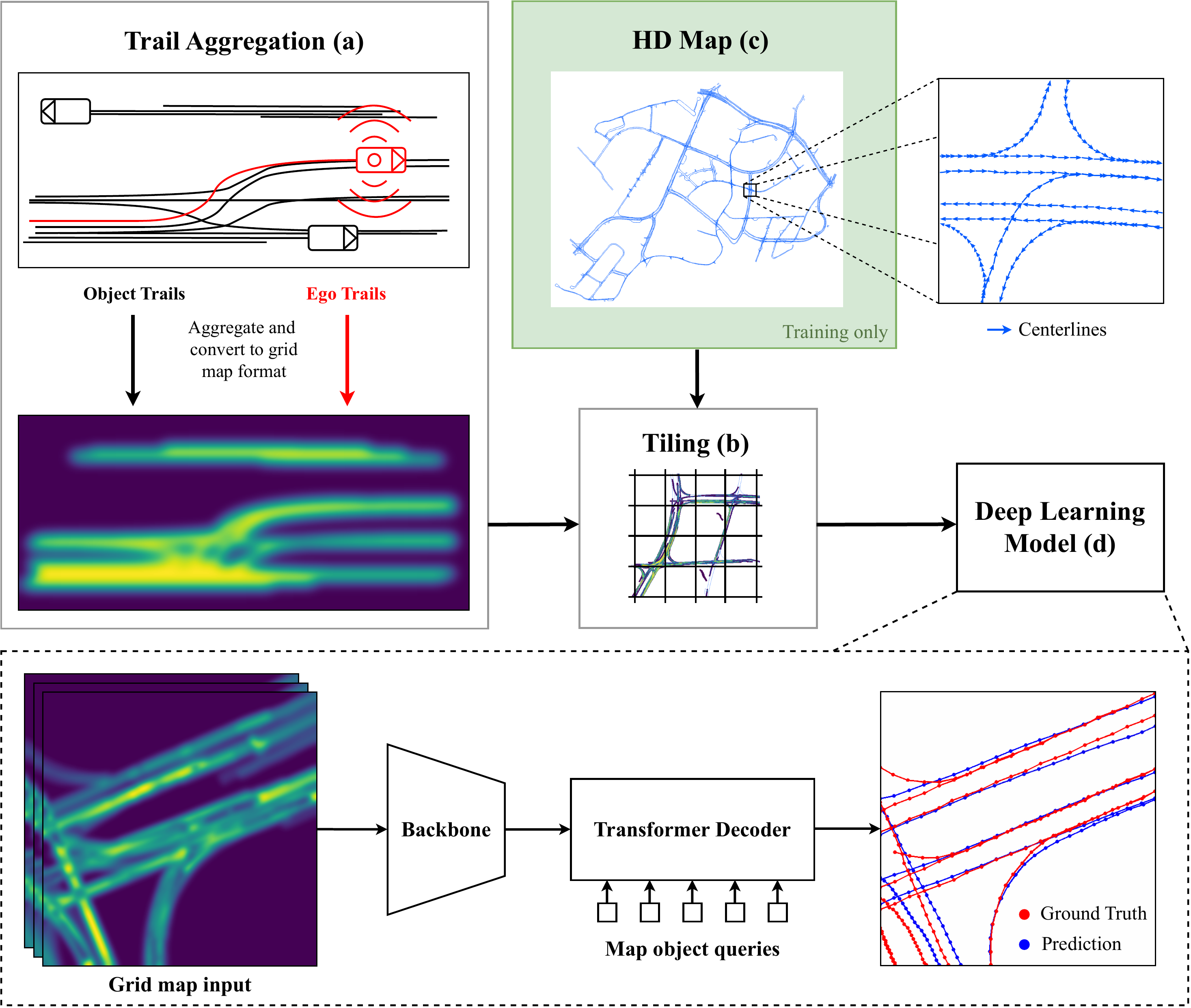}
\caption{The overall architecture of the trail-based mapping approach. Starting with the trail aggregation (a), object and ego trails are converted into a grid map format. The covered area is divided into tiles (b) and for training the ground truth is obtained from a centerline HD map (c). The resulting samples are then fed into a deep learning model (d) that, once trained, infers tile-wise centerline predictions for the input trail grid maps.}
\label{fig:overview}
\end{figure*}



\vspace{7mm} 

\noindent \textbf{Trail and crowdsourced mapping}:{\enspace\;}Trail-based approaches utilize recordings of vehicle trajectories for map creation. In general, most early approaches aggregated GPS trails of multiple recording vehicles and extracted the road network structure based on rule-based methods like trace-merging, k-means clustering, or kernel density estimation \cite{biagioni_inferring_2012}.

Building on these foundational techniques, more advanced trail-based methods have been developed for lane mapping and road geometry extraction. One approach involved generating BEV orthographic images by fusing GPS, INS, and visual odometry data, then deriving lane graphs using a rule-based method incorporating the Random Sample Consensus algorithm \cite{guo_low-cost_2016}. Another method inferred lane geometries from Radar and camera probe data by measuring distances to stationary objects and applying k-means clustering to identify lane markings and widths \cite{massow_deriving_2016}.

A specialized highway mapping approach used an iterative Graph SLAM method, aligning data from odometry, GNSS, and traffic sign measurements, followed by road boundary measurements for refinement \cite{doer_hd_2020}. A key challenge in offline mapping is maintaining up-to-date maps, which Kim et al. addressed by incorporating uncertainty estimation to identify reliable landmark observations \cite{kim_hd_2021}. Another method introduced a three-stage pipeline for detecting, updating, and integrating map changes \cite{pannen_how_2020}.

Integrating 3D object tracking modules allows for the inclusion of additional trails from observed traffic participants, expanding the available trail data. One mapping approach involved dividing aggregated trails into tiles and constructing a density map to visualize detection frequencies per pixel \cite{colling_hd_2022}. The method further separated density maps by driving direction and used a rule-based algorithm to extract lane-level centerlines.


A specialized approach for intersections estimated lane geometries using detected road user trajectories. It employed a two-stage Markov chain Monte Carlo sampling method \cite{meyer_anytime_2019}, later refined with a non-linear least squares formulation for greater accuracy \cite{meyer_fast_2020}. The system could process data from cameras, Radar, or Lidar but was primarily evaluated using simulated vehicle data. Similarly, FlowMap \cite{ding_flowmap_2023} introduced a framework for deriving human-like driving paths in unmarked road environments by clustering entry and exit points of recorded trajectories. An extended version incorporated traffic flow data to generate HD map patches for intersections \cite{qin_traffic_2023}.



A more recent approach automated lane model generation by leveraging sparse vehicle observations and a deep learning-based architecture to predict lane pairs and their connectivity \cite{mink_lmt-net_2024}. While the model demonstrated promising results, its evaluation was limited to internal datasets, restricting generalizability.

Among the aforementioned methods, only one specifically addresses centerline extraction from trail data using state-of-the-art deep learning models. In our approach, we integrate recent advancements from online mapping into an offline mapping pipeline. Therefore, we adopt a BEV tile-based structure, capturing information from aggregated pre-recorded data, and modify the model architecture accordingly. Once trained, our deep learning-based model enables tile-wise map updates, increasing flexibility and scalability compared to existing methods.
\section{Methodology}

\begin{figure*}[t] 
    \centering
    \subfloat[Visualization of an example trail grid map sample without yaw angle separation.]{\includegraphics[width=0.25\textwidth,valign=t]{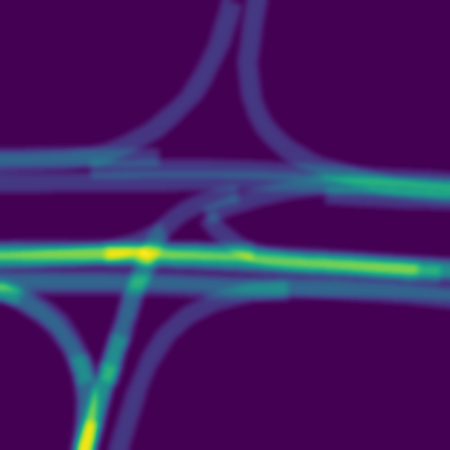}\label{fig:heatmap_sum}}
    \quad
    \subfloat[Visualization the trail grid map sample from \cref{fig:heatmap_sum} separated into $n = 6$ different bins by yaw angle.]{\includegraphics[width=0.38\textwidth,valign=t]{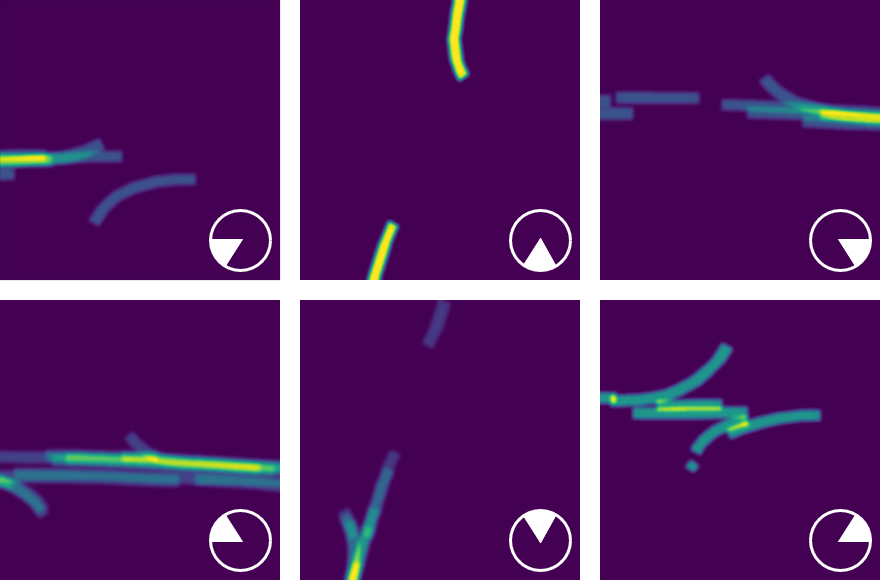}\label{fig:separated_heatmap}}
    \quad
    \subfloat[Visualization of the average speed profile for the example sample from \cref{fig:heatmap_sum}.]{\includegraphics[width=0.25\textwidth,valign=t]{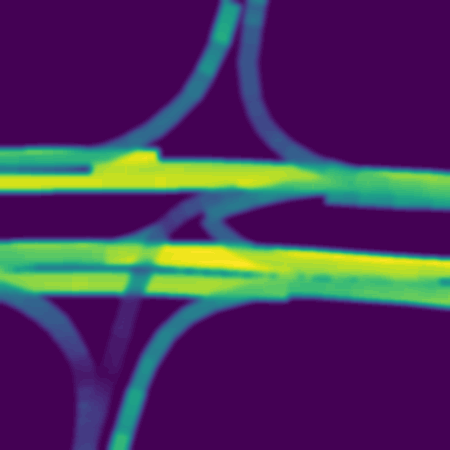}\label{fig:speed_profile}}
    \caption{Visualization of an example sample from the nuScenes dataset.}
    \label{fig:heatmaps}
\end{figure*}

In the following section, the overall method of our trail-based mapping approach is outlined. Firstly, the input and ground truth representation are described, followed by the data augmentation for training, our model architecture and the output representation.

\subsection{Input and ground truth representation}\label{sec:input_and_gt} 


The first step of our proposed method is trail aggregation (cf. \cref{fig:overview}a). Vehicles equipped with self-localization and 3D perception sensors record driving data, which is subsequently processed to integrate localization information with an object detection and tracking algorithm. This results in two distinct types of trails: Ego trails representing the trajectory of the ego vehicle during the recording period and object trails depicting the perceived motion paths of other vehicles in traffic. A trail $t$ is defined by a series of cuboid center points $c_i, \, i \in [1, p]$, forming a linearized path $t = \{c_1, c_2, ..., c_p\} $. The trail width $w_t$ corresponds to the width of the respective vehicle $w_{\textit{veh}}$.

The trails are then mapped onto a global coordinate system. To serve as input for a neural network, the trails are converted into a grid-based map representation. This process involves dividing the covered area into discrete tiles (cf. \cref{fig:overview}b). For a single tile, the real world area $w_{\textit{glob}} \times h_{\textit{glob}}$ with a resolution of $r$ results in a representative grid map $G$ of size $w_{\textit{grid}} \times h_{\textit{grid}} :=\frac{w_{\textit{glob}}}{r} \times \frac{h_{\textit{glob}}}{r}$.

$G$ is separated into $n$ different bins based on the current yaw angle $\varphi$ of each trail segment, which enables clear separation of directions at intersections and distinction of opposing lanes (cf. \cref{fig:separated_heatmap}). The yaw angle of a trail segment is defined as $\varphi \in (-\pi, \pi]$ in the global coordinate system, resulting in an interval of $I_j$ per bin:

\begin{equation}
    I_{j} = \biggl( -\pi + (j-1) \frac{2\pi}{n}, -\pi + j \frac{2\pi}{n}\biggr] , \quad j \in [1, n].
\end{equation}

Each bin corresponds to an individual channel $ C_{\textit{dir}, j} \in \mathbb{R}^{w \times h}$ of $G$. For each trail segment, the values of the individual direction channel $C_{dir, j}$ are incremented by $1$ for each containing grid cell.

Additionally, the average speed for each grid cell is computed and incorporated as an additional input channel $C_{\textit{speed}} \in \mathbb{R}^{w \times h} $ (cf. \cref{fig:speed_profile}). 

The resulting definition of $G$ is given by:

\begin{equation}
    G \in \mathbb{R}^{w \times h \times (n+1)}.
\end{equation}

To ease the detection of sensible features by the neural network, Gaussian smoothing is applied to each input matrix channel. The kernel for the smoothing operation is characterized by the standard deviation $\sigma$ of the normal distribution and determines the degree of smoothing applied to the input data.




To train our model, ground truth information is required in addition to the trail-based input. In our approach, we derive the ground truth from an HD centerline map, ensuring alignment with the areas covered by the recorded trails (cf. \cref{fig:overview}c). Since the neural network is expected to predict centerline directions, the HD map must provide directional information for each centerline. To generate the training data, we extract centerlines from the HD map that correspond to regions with nonzero trail density while explicitly filtering out centerlines that lack associated trails. This filtering process is performed by matching the yaw angles of the centerlines and trails to ensure consistency. From the remaining centerline graph, we derive all drivable paths following the approach of \cite{liao_lane_2024}. We represent each path by a fixed number of ordered points $m$.

\subsection{Data augmentation} \label{sec:augmentation}

In addition to the grid-based tiling shown in \cref{fig:overview}b, we apply a sample augmentation method to enhance training diversity. Instead of partitioning the regions containing trails and ground truth data into fixed grid tiles, we randomly select center points within these areas to define augmented tiles. To introduce further variation, we assign the rotation to each tile randomly and we only select a subset of trails within the tile area, ensuring exposure to different trail distributions during training.

\subsection{Model architecture and output}

The model architecture builds upon MapTRv2 \cite{liao_maptrv2_2023}, utilizing the same DETR-inspired query-based transformer decoder. However, we introduce modifications to the backbone to better accommodate our input representation. We replace the original multi-view camera and Lidar processing pipeline with a single ResNet50 backbone for feature extraction. This change is motivated by the fact that our input consists of a multi-channel matrix. Several auxiliary task heads, including depth prediction, BEV segmentation, and perspective-view segmentation, are removed, as they are not applicable to our setting. The classification head is also omitted, as our model focuses exclusively on centerline prediction.

The maximum number of predicted centerlines in the output of the model is a tunable hyperparameter that correlates with the amount of queries in the transformer decoder. Each prediction consists of the same fixed number of points $m$ compared to the ground truth centerlines. An example of a prediction output and corresponding ground truth is shown in the bottom right of \cref{fig:overview}. The order of the coordinates in the prediction indicates the direction of the centerline.

The loss of the model is based on instance and point-level matching as described in \cite{liao_maptrv2_2023}, with only a single possible permutation per centerline since its direction is fixed. We employ the same one-to-one and one-to-many set prediction loss, but omit the focal loss since there is no classification head.
\section{Experiments}

\begin{table*}[t]
\centering
\caption{Validation split performance of MapTRv2 in comparison to our approach (TrailTR) for different training and validation splits.}
\setlength{\tabcolsep}{16pt}
    \label{tab:gen_areas_results}
    \begin{tabularx}{0.97\textwidth}{@{} l *{2}{l} >{\centering\arraybackslash}p{2.8cm} @{}}
        \toprule
        Model & Training Data & Validation Data & Validation Split AP \\ 
        \midrule
        MapTRv2    & nuScenes, original train split    & nuScenes, original val split    & 53.3 \\ 
        TrailTR    & nuScenes, original train split    & nuScenes, original val split    & \textbf{56.4} \\ 
        \addlinespace
        \hline
        \addlinespace
        MapTRv2    & nuScenes, geographical train split    & nuScenes, geographical val split    & 30.5 \\ 
         & & \: + trail area masking & 27.2 \\
        TrailTR    & nuScenes, geographical train split    & nuScenes, geographical val split    & \textbf{44.6} \\ 
        \addlinespace
        \hline
        \addlinespace
        TrailTR    & nuPlan, original train split    & nuPlan, original val split    & \textbf{70.3} \\ 
        \addlinespace
        \hline
        \addlinespace
        TrailTR    & nuScenes, geographical train split    & nuPlan, original val split    & 34.0 \\ 
        TrailTR    & nuPlan, original train split    & nuScenes, geographical val split    & 35.0 \\ 
        TrailTR    & \makecell{nuScenes, geographical train split \\ + nuPlan, original train split}    & nuPlan, original val split    & 69.7 \\ 
        TrailTR    & \makecell{nuScenes, geographical train split \\ + nuPlan, original train split}    & nuScenes, geographical val split    & 44.3 \\ 
        \bottomrule
    \end{tabularx}
\end{table*}

\begin{table*}[t]
\centering
\caption{Validation split performance of MapTRv2 in comparison to our approach (TrailTR) on geographical nuScenes training and validation splits with different input data foundation modalities.}
\setlength{\tabcolsep}{9pt}
    \label{tab:gen_input_mod_results}
    \begin{tabularx}{\textwidth}{@{} l *{3}{l} >{\centering\arraybackslash}p{2.8cm} @{}}
        \toprule
        Model & Input Data Foundation & Training Data & Validation Data & Validation Split AP \\ 
        \midrule
        MapTRv2    & Camera     & nuScenes, geo. train split    & nuScenes, geo. val split    & 30.5 \\
        TrailTR    & Ground truth & nuScenes, geo. train split    & nuScenes, geo. val split    & 44.6 \\
        TrailTR    & Camera + Lidar & nuScenes, geo. train split    & nuScenes, geo. val split    & 43.7 \\ 
        \addlinespace
        \hline
        \addlinespace
        TrailTR    & Radar & \makecell{nuScenes, geo. train split \\ + nuPlan, original train split}    & closed Las Vegas dataset    & 30.0 \\ 
        \bottomrule
    \end{tabularx}
\end{table*}

In the following section, we present experiments of our trail-based mapping approach. We first describe the datasets. Next, we compare our method's performance against state-of-the-art online mapping approaches. Finally, we conduct ablation studies to assess the impact of input variations on performance.

\subsection{Datasets}

To ensure reproducibility and comparability with existing approaches, we primarily rely on public datasets for training and evaluation. These datasets must contain both trail information and ground truth centerlines to be suitable for our approach. Trail information can be given by recorded GPS poses of the ego vehicle, combined with dynamic object data that includes tracking over time. Ground truth centerlines must be embedded within a global HD map, covering the entire trail region. Before a dataset can be used for training, the sample generation process (cf. \cref{fig:overview}a-c) must be applied to extract the necessary input representations.

In online mapping research, the most widely used datasets are nuScenes \cite{caesar_nuscenes_2020} and Argoverse2 \cite{wilson_argoverse_2023}. Despite containing only 15 hours of driving data, nuScenes provides a detailed HD map for each included city \cite{caesar_nuscenes_2020}. In Argoverse2, the HD map is divided into local, scenario-specific tiles, which renders it incompatible with our proposed tiling and augmentation pipeline. In contrast, the nuPlan dataset \cite{caesar_nuplan_2022} provides city-wide HD maps and significantly larger amounts of trail data, with 1,500 hours of recorded driving data. 

For comparability with online mapping approaches, we train and evaluate our method on the nuScenes dataset. To assess transferability and generalization, we further validate our approach on the nuPlan dataset as well as additional closed data sources. Since the selected datasets do not provide trail data for every centerline element in the HD map, each generated tile is refined according to the procedure described in \cref{sec:input_and_gt}.


\subsection{Comparison to online HD mapping approaches}

\begin{figure*}[t] 
    \centering
    \subfloat[]{\includegraphics[width=0.29\textwidth,valign=t]{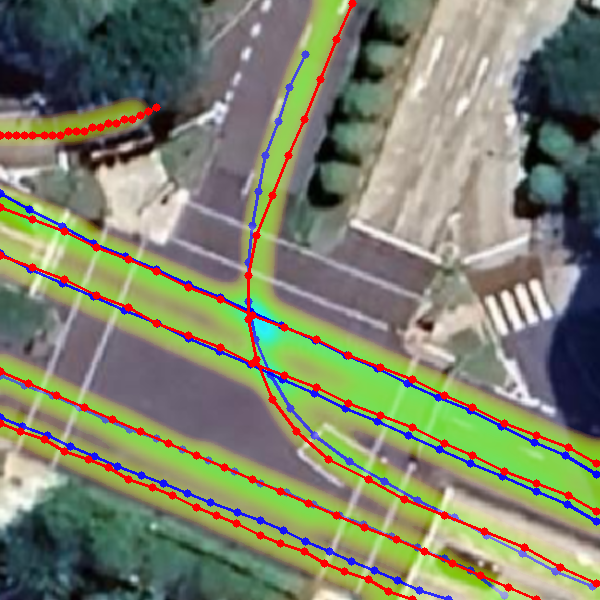}\label{fig:nusc_example_1}}
    \quad
    \subfloat[]{\includegraphics[width=0.29\textwidth,valign=t]{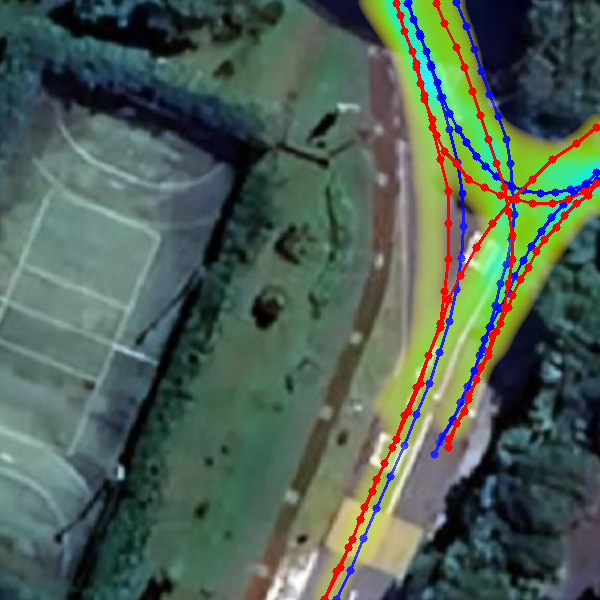}\label{fig:nusc_example_2}}
    \quad
    \subfloat[]{\includegraphics[width=0.29\textwidth,valign=t]{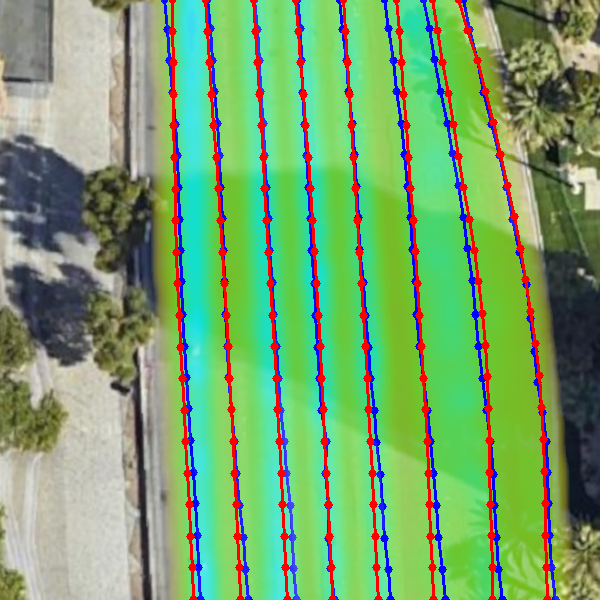}\label{fig:nuplan_example_1}}
    \\
    \subfloat[]{\includegraphics[width=0.29\textwidth,valign=t]{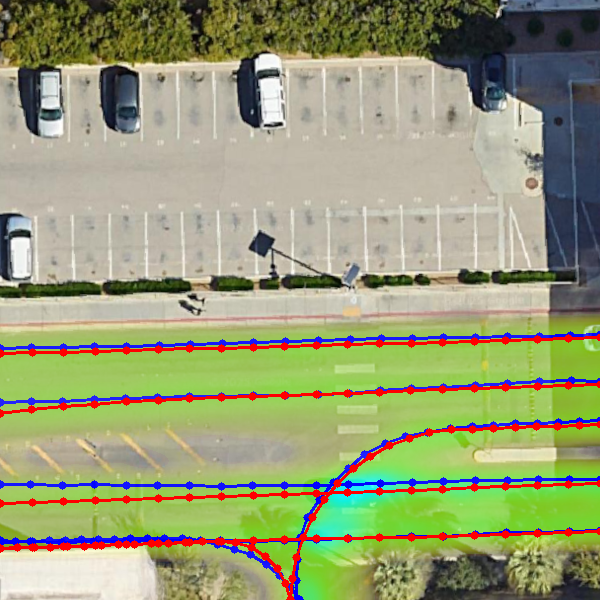}\label{fig:nuplan_example_2}}
    \quad
    \subfloat[]{\includegraphics[width=0.29\textwidth,valign=t]{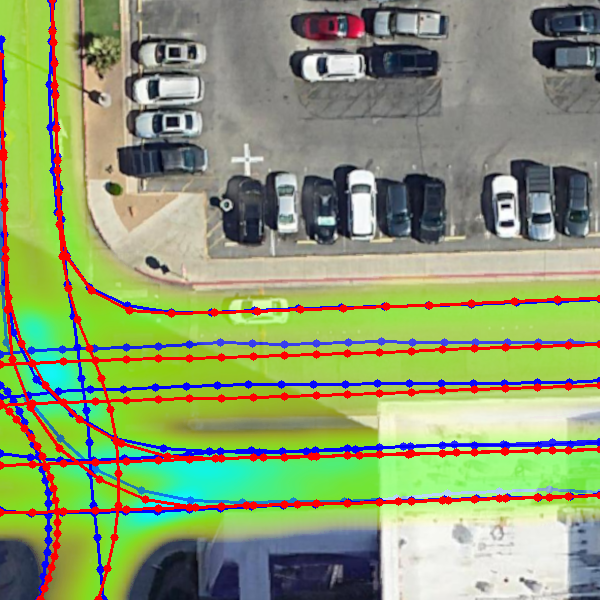}\label{fig:nuplan_example_3}}
    \quad
    \subfloat[]{\includegraphics[width=0.29\textwidth,valign=t]{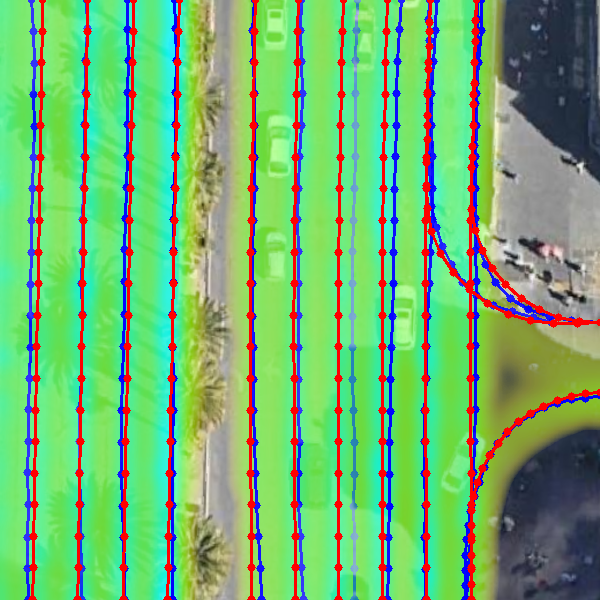}\label{fig:nuplan_example_4}}
    \caption{Visualization of ground truth (red), model predictions (blue) and its input grid map on an aerial image. The samples are from the nuScenes geograhical validation split (a-b) and nuPlan validation split (c-f). The samples from the nuScenes dataset lack sufficient amounts of trails to depict the entirety of lanes in the map tile, so the ground truth was refined as described in \cref{sec:input_and_gt}.}
    \label{fig:example_samples}
\end{figure*}


\noindent \textbf{Metric}:{\enspace\;}For evaluation, we employ the same metric as in most online mapping approaches. We calculate the average precision ($AP$) for all samples by utilizing the Chamfer distance $D_{\textit{Chamfer}}$ as a threshold to determine the positive matches between prediction and ground truth. The $AP$ is calculated for each of the three thresholds $\tau \in T, T = \{0.5, 1.0, 1.5\}$ individually and then averaged across all to determine the final $AP$ metric used for comparison:

\begin{equation}
    AP = \frac{1}{|T|} \sum_{\tau \in T} AP_{\tau}.
\end{equation}



\vspace{5mm} 

\noindent \textbf{Generalization to new areas}:{\enspace\;}A recent study highlighted limitations in the original training, validation, and testing splits of the nuScenes dataset, particularly regarding geographical overlaps between these splits \cite{lilja_localization_2024}. The authors show, that most state-of-the-art online mapping models have been trained and evaluated on the original splits and show limited generalization when applied to geographically disjoint splits. To assess whether this trend also affects centerline detection in online mapping, we trained MapTRv2 \cite{liao_maptrv2_2023} using both the original nuScenes split and geographically disjoint near extrapolation nuScenes split \cite{lilja_localization_2024}. The results of this evaluation are presented in \cref{tab:gen_areas_results}.

For training and evaluation of our trail-based approach, we delete ground truth data for areas with no trails. In order to compare the results for MapTRv2 with our approach, we employ masking for the respective validation splits. Thereby, we omit the Chamfer distances $D_{j, \textit{Chamfer}}$ for areas for which no trails exist during the calculation of the $AP$. This way, it is ensured that the areas without trails do not decrease the performance of the MapTRv2 results (e.g. because of increased complexity inside the masked area) in comparison to the trail-based approach. The results for the centerline $AP$ with masking are shown in \cref{tab:gen_areas_results}.

Our trail-based approach outperforms MapTRv2 on both the original (56.4 AP, +5.8 \%) and geographically disjoint near extrapolation (44.6 AP, +46.2 \%) nuScenes splits, demonstrating superior generalization across different data distributions. Additionally, our approach benefits from larger and more diverse training datasets. This is evident from the performance improvement observed when training and evaluating on the original nuPlan training and validation splits, which contain a significantly higher number of trails compared to nuScenes and lead to a performance of 70.3 AP on the validation split. The predictions for exemplary samples from the nuScenes geographical validation split and the nuPlan validation split are visualized in \cref{fig:example_samples}. When cross-evaluating the two datasets, a performance decrease can be observed, which may be caused by the difference in trail amounts per tile between nuScenes and nuPlan. By training on both datasets, this effect can be mitigated and the validation split performance on both datasets reaches similar performance compared to trainings on the individual datasets.


\vspace{5mm} 

\noindent \textbf{Generalization to different sensor setups}:{\enspace\;}To avoid introducing additional noise during training due to sensor imperfections, we utilized the ground truth object tracking annotations provided by the nuScenes and nuPlan datasets to extract trail data for our previously listed experiments. To verify the sensor-agnostic properties of our approach, we additionally extracted sensor-based trail data using the state-of-the-art object detection and tracking algorithm FocalFormer3D \cite{chen_focalformer3d_2023}. The results, presented in \cref{tab:gen_input_mod_results}, indicate a minimal performance decrease, as the general shape of the grid tiles is preserved through sufficiently accurate object tracking. Furthermore, we applied our approach to a closed Las Vegas dataset, using Radar-based detections as the foundation for trail data. The corresponding results are also listed in \cref{tab:gen_input_mod_results}.

\subsection{Ablation Studies}\label{sec:ablation}

To further optimize the centerline extraction, we provide ablation studies regarding modifications to the model input. 

\vspace{5mm} 

\noindent \textbf{Gaussian smoothing}:{\enspace\;}To ease the detection of sensible features in the input, we employ Gaussian smoothing on each individual input channel. \Cref{tab:ablation_studies_smoothing} presents the results for various standard deviations $\sigma$ of the Gaussian smoothing kernel. It appears that the optimal value is $\sigma = 2$, as it effectively unifies visual features across different input samples while preserving the overall shape of trail densities.

\begin{table}[t]
\centering
\caption{Different smoothing settings and the corresponding performance on the geographical nuScenes validation split.}
\setlength{\tabcolsep}{10pt}
    \label{tab:ablation_studies_smoothing}
    \begin{tabularx}{0.47\textwidth}{@{} l >{\centering\arraybackslash}p{3.95cm} @{}}
        \toprule
        Smoothing Settings & Validation Split AP \\ 
        \midrule
        No smoothing    & 44.1 \\
        Gaussian smoothing, $\sigma$ = 1    & 44.1 \\
        \textbf{Gaussian smoothing, \boldmath$\sigma$ = 2}    & \textbf{44.6} \\
        Gaussian smoothing, $\sigma$ = 4    & 41.5 \\
        Gaussian smoothing, $\sigma$ = 8    & 38.9 \\
        \bottomrule
    \end{tabularx}
\end{table}

\begin{table}[t]
\centering
\caption{Different channel combinations and binning strategies with the corresponding performance on the geographical nuScenes validation split.}
\setlength{\tabcolsep}{0.1pt}
    \label{tab:ablation_studies_channel_variation}
    \begin{tabularx}{0.47\textwidth}{@{} l >{\centering\arraybackslash}p{3.3cm} @{}}
        \toprule
        Channel Settings & Validation Split AP \\ 
        \midrule
        Single trail grid channel    & 9.4 \\
        \: \: \: + speed profile    & 36.9 \\
        $n$ = 4 channel bins + speed profile    & 44.1 \\
        \textbf{\boldmath$n$ = 6 channel bins + speed profile}    & \textbf{44.6} \\
        $n$ = 12 channel bins + speed profile    & 43.4 \\
        $n$ = 36 channel bins + speed profile    & 43.2 \\
        \bottomrule
    \end{tabularx}
\end{table}


\vspace{5mm} 

\noindent \textbf{Channel variations and speed profile}:{\enspace\;}The network is designed to predict directed centerlines. Introducing directional bins helps distinguish individual directions, providing additional information to the model. Furthermore, incorporating the average speed profile offers insights into lane structure, as speed variations often correlate with specific road layouts. In \cref{tab:ablation_studies_channel_variation}, we present the performance differences resulting from the inclusion of the speed profile and the use of different binning strategies. The speed profile and direction bin separation increase the input complexity while providing additional information, therefore we expect that an optimal balance exists. We experimentally verified our assumptions and found an optimum at $n = 6$.



\vspace{5mm} 

\noindent \textbf{Augmentation}:{\enspace\;}As described in \cref{sec:augmentation}, we applied an augmentation technique to enhance training diversity. To assess its impact on performance, we additionally trained the model without augmented tiles and report the results in \cref{tab:ablation_studies_augmentation}. As expected, the additional training samples increased the performance on the validation split (+77.0 \%).

\begin{table}[t]
\centering
\caption{Training with and without augmented samples from the nuScenes geographical training split and the corresponding performance on the geographical nuScenes validation split.}
\setlength{\tabcolsep}{0.1pt}
    \label{tab:ablation_studies_augmentation}
    \begin{tabularx}{0.47\textwidth}{@{} l >{\centering\arraybackslash}p{3.0cm} @{}}
        \toprule
        Training Samples & Validation Split AP \\ 
        \midrule
        Training without augmented samples    & 25.2 \\
        \textbf{Training with augmented samples}    & \textbf{44.6} \\
        \bottomrule
    \end{tabularx}
\end{table}




\section{Conclusion}

In this work, we introduced a novel trail-based mapping approach that extracts high-definition driving maps by aggregating trajectories from both ego vehicles and other traffic participants. Our method leverages deep learning to convert trail data into a map representation, enabling tile-wise global map construction. Unlike traditional offline HD maps, our approach is sensor-agnostic and scalable, allowing for continuous map updates without dependence on predefined static datasets.

Through extensive experiments, we demonstrated that our method outperforms state-of-the-art online mapping models, particularly in terms of generalization to new areas. Our approach remains robust across various sensor setups, with minimal performance degradation when using sensor-based trail extraction. Additionally, we showed that augmenting training data with diverse trail distributions further enhances model performance.

While this work focuses on centerline extraction, the methodology can be extended to extract additional road elements, such as lane boundaries, pedestrian crossings, and traffic signs, by incorporating corresponding sensor data. Furthermore, integrating our method into a comprehensive map generation framework could enable the construction of fully connected lane graphs, further bridging the gap between online and offline mapping approaches.

\section*{Acknowledgements}
This work is a result of the joint research project STADT:up (19A22006B) The project is supported by the German Federal Ministry for Economic Affairs and Climate Action (BMWK), based on a decision of the German Bundestag. The authors are solely responsible for the content of this publication.
{
    \small
    \bibliographystyle{ieeenat_fullname}
    \bibliography{main}
}


\end{document}